# Identifying Populist Paragraphs in Text: A machine-learning approach

**Authors:** *Jogilė Ulinskaitė[1] and Lukas Pukelis[2]*

**Abstract:** *In this paper we present an approach to develop a text-classification model which would be able to identify populist content in text. The developed BERT-based model is largely successful in identifying populist content in text and produces only a negligible amount of False Negatives, which makes it well-suited as a content analysis automation tool, which shortlists potentially relevant content for human validation.*

**Introduction**

This paper presents our attempt to develop a machine-learning (ML) model to detect populist content in text. If successful, this model could benefit many researchers by automating the most resource-intensive part of the research and enabling more extensive and more ambitious research projects. This methodological improvement could enable more detailed and broader comparative analyses, leading to a better understanding of populism.

However, as attractive as this may seem, there are some critical challenges to overcome to develop such a model. First, the term "populism" has received much attention in academia for several years now, which has led to a proliferation of different definitions and, in many cases, a vague operationalization and concept-stretching (Pappas 2016). In order to develop an operational definition of 'populism', a comprehensive literature analysis is necessary. Furthermore, this definition should focus on the intrinsic characteristics of populism and not depend on the national context, the register of text, or the author's ideological position. The second challenge is to assemble a training dataset for machine-learning models that is large and diverse enough to allow the developed model to perform well with diverse previously unseen data. Finally, the third challenge is to validate the performance of the developed model in a way that provides a realistic understanding of how the model would perform "in the wild," i.e. on new data that might differ from the training data in a significant number of ways.

In our approach, we define populism primarily as a discursive strategy that actors across the ideological spectrum can employ. We see "populism" as composed of two distinct components - people-centrism (referring to "the people" as a single entity with homogeneous interests) and anti-elitism (a sentiment that the current governing elites are corrupt and act against the interests of "the people"). These two components, although sometimes appearing together, are distinct and have been coded separately in our analysis. In addition, we developed two sets of ML models to detect these two dimensions of populism. To train the models, we have developed a new dataset based on the established data sources, where each paragraph of text is coded as containing or not containing people-centric or anti-elitist sentiment.

[1] Lecturer Vilnius University Institute of International Relations and Political Science (VU IIRPS) (jogile.ulinskaite@tspmi.vu.lt)

[2] Data Scientist Public Policy and Management Institute (PPMI) (Lukas.pukelis@ppmi.lt)

To validate the model's performance, we prepared a separate dataset by manually coding the 2016 and 2020 election manifestos of Lithuanian political parties. We have carried out the validation to simulate a real-life scenario where a researcher uses the ML model for a specific research project. As we have not used data from Lithuania in the original training set, this reduced the risk of contamination when the data "testing" the model is not new but somehow appeared in the training set. To further reduce the risk, we split the Lithuanian dataset into two parts and used one part as a "test" during model development and the other as a "hold-out" once model development was complete.

The developed model performed reasonably well on the validation dataset (accuracies of 0.86 and 0.95 for people-centrism and anti-elitism, respectively). It had a slight tendency to over-predict (generate false positives), which is acceptable as it is designed to act as an automation aide for researchers, with human coders checking and validating its predictions.

The paper structure is as follows: the first part presents an overview of existing research and the operational definition of populism used in this paper. The second part describes the data and methodology, and the third part presents the results of the model validation.

## 1. Overview of existing research

Research on populism has started with, and for a long time, dominated by, in-depth analyses of specific cases of populism (Grabow & Hartleb, 2013, Mudde & Kaltwasser, 2012). Recent research seems to shift focus on broader scale comparative analysis both country-wise, period-wise, and source-wise. Classical content analysis, started by J. Jagers and S. Walgrave (2007), is still one of the most widely used populist discourse analysis methods. With slight differences between the methods, researchers most often comparethe proportion of populist content by coding specific excerpts of texts such as a paragraph (Rooduijn & Pauwels, 2011, Rooduijn, 2014, Pauwels & Rooduijn, 2015, Rooduijn & Akkerman, 2017), a statement (Ernst et al., 2017, Manucci & Weber, 2017, Ernst et al., 2019, Bernhard & Kriesi, 2019), an issue-specific claim (Bernhard et al., 2015), a sentence (Vasilopoulou et al., 2014) or a quasi-sentence (March, 2018). Researchers frequently attempt to maintain the validity of the classical content analysis and make the process easier by adding semi-automation tools (Caiani & Graziano, 2016, Ernst et al., 2017, Wettstein et al., 2018, Ernst et al., 2019).

Since classical content analysis is very time- and labour-consuming, more extensive comparative studies involve automated methods such as the dictionary-based approach (Pauwels, 2011). Even though the computer-based coding method's validity is somewhat lower than the classical content analysis (Rooduijn & Pauwels, 2011), both approaches generate reasonably valid results (Storz & Bernauer, 2018). The dictionary-based approach is extensively used to analyze both media content (Hameleers & Vliegenthart, 2020, Gründl, 2020) and party-generated data (Storz & Bernauer, 2018, Elçi, 2019, Payá, 2019). Further developments of the method (Bonikowski & Gidron, 2016) and manual check of the text excerpts (Pauwels, 2017) have been suggested to improve the validity of the dictionary-based approach.

Holistic grading is another approach specifically developed to make manual coding more efficient. The method combines the benefits of classical content analysis (holistic approach, human interpretation) and dictionary-based approach (ability to compare large amounts of data). The whole text (usually a speech) is coded by human coders using an explicit rubric (Hawkins, 2009). The approach enabled researchers to develop the Global Populism Database consisting of various political texts (Hawkins et al., 2019).

Finally, at least several different expert-based populist databases have been established in recent years: Populism and Political Parties Expert Survey (POPPA), The PopuList, The Global Party Survey and Timbro Authoritarian Populism Index. They categorize populist political actors and provide a more comprehensive perception of the populist phenomena across different regions and time-frames.

Despite various methodological improvements and developments, automated textual analysis has not yet gained momentum. The exception is an attempt by Hawkins and Silva (2018) to use elastic-net regression for the supervised classification of party manifestos. Their results suggest that the model can identify very populist manifestos and very not-populist documents but does not perform very well on the documents in-between. They conclude that using more training data could improve results. We also suggest that dividing and hand-coding shorter excerpts of manifestos (paragraphs) could improve the model.

In recent years, artificial neuron network models have demonstrated outstanding results in many spheres of application. Arguably, since 2018 the biggest progress has been made in the area of natural language processing, where these models have been applied to a number of natural language understanding and text-classification tasks. Given the magnitude of these improvements, it is prudent to expect that similar techniques could be also applied to improve the classification of the populist text.

## 2. Methods

To develop a machine learning model to recognize populist content, we have employed a standard machine-learning workflow: first, we collected and prepared a training dataset used to develop a machine-learning model (more precisely, an ensemble of models). The performance of the models was tested using a small set of manually coded Lithuanian political party manifestos. Finally, the performance of the trained model was validated using a more extensive, separately coded dataset of all the Lithuanian political party manifestos from the 2016 and 2020 parliamentary elections. This train-test-holdout approach was chosen because the more commonly used train-test approach can often lead to unintended overfitting, as optimizing the model for its performance on the test dataset may still introduce some contamination effects that degrade the overall performance of the model (Roelofs, et al. 2019).

## 2.1. Train and Test Datasets

### 2.1.1. Defining and operationalizing populism

Theoretical studies of populism usually conceptualise it as an ideational phenomenon: as a set of ideas lacking fundamental values (Taggart, 2002), an undeveloped thin-centred ideology with its specific concepts (Canovan, 2002), a recurrent feature of modern politics using specific themes (Arditi, 2003, 2007). Cas Mudde (2004, pp.543) has formulated the most often used definition: a thin-centred ideology "that considers society to be ultimately separated into two homogeneous and antagonistic groups, 'the pure people' versus 'the corrupt elite', and which argues that politics should be an expression of the volonté générale (general will) of the people". Following the theoretical conceptualizations, we claim that populism as an ideology is reflected in the discourse (Pauwells, 2011). We consider it an attribute of a text rather than a feature of a politician (Rooduijn, 2014).

For coding, we follow the instructions suggested by (Rooduijn & Pauwels, 2011). The coding unit is a paragraph, as it allows to distinguish between different arguments (Pauwels, 2011) and is a sufficiently long passage of text to elaborate on people-centrism and anti-elitism (Rooduijn, de Lange and van der Brug, 2014). Populism is defined as a thin-centred ideology having two main elements: people-centrism and anti-elitism. A paragraph is coded as people-centrist if it refers to a general category of the people as a homogeneous unit having favourable properties. It is important to distinguish when the author of the text refers to individuals, distinct groups of society (e.g. women, children, pensioners) or society in general. Only when a paragraph refers to people, society, citizens, nation, we code it as people-centrist. In addition, we code a paragraph as people-centrist when singular words such as a person or a citizen in the text refer not to a specific individual but an individual representing the whole. We identify anti-elitism if a paragraph refers to the elite as a homogeneous group having negative properties. A paragraph is coded as anti-elitist when the criticism is generalised to the government, politicians, bureaucracy, oligarchy, financial, cultural or academic elites. When criticism of the elite refers to particular political parties or officeholders, we do not consider it anti-elitist.

### 2.1.2. Data sources for manual coding

We started with pilot coding of populist paragraphs in the most populist speeches from Team Populism Datasets. We first coded texts that were anchor texts for coders in the Global Populism Database (2019). We then proceeded with populist speeches, aiming for various regions, document types, left and right populism. Finally, we coded populist paragraphs in the manifestos of often-recognized populist political parties: Manifesto for European Elections of Die Linke (2014), The Political Programme of the Alternative for Germany (2017), Law and Justice Party Program (2014).

Since the number of paragraphs containing both elements of populism (people-centrism and anti-elitism) in those very populist texts was surprisingly low, we decided to code the two elements of populism separately. We require the two elements to be present in the text to be considered populist, but they do not necessarily have to occur in one sentence, claim, or paragraph. For example, people-centrism can be more prevalent in the first half of the manifesto and anti-elitism in the second. We approach populism as a gradational phenomenon, meaning that the populism level of a particular text can be interpreted meaningfully only compared to other textual material.

The second stage of coding involved coding manifestos of parliamentary parties identified as populist by *The PopuList* database. We have manually coded 30 manifestos retrieved either from the websites of the political parties or from the Comparative Manifesto Project database. Finally, to increase the number of anti-elitist paragraphs, we have searched for anti-elitist parties according to the Chapel Hill Survey (ANTIELITE_SALIENCE = salience of anti-establishment and anti-elite rhetoric). We have added anti-elitist paragraphs to our dataset from the political manifestos of Die Tierschutzpartei (Germany, 2019) and Zivi zid (Croatia, 2015).

**Non-Populist Texts**
We aimed to develop a model that could provide satisfactory performance in various contexts and texts in various registers (long-form party election manifestos, short manifesto summaries, media articles, etc.). For this reason, we have decided to add some non-populist texts from the public domain to the training set to introduce more diversity and prevent overfitting. In selecting the texts, we aimed to include documents with specific characteristics similar to those of populist texts but were clearly non-populist. For instance, we added the US Constitution as an example of statist discourse and Tolstoy's "War and Peace" as an example of more emotionally elevated speech.

**Test Dataset**
The test dataset consisted of manually coded paragraphs from the manifestos of the two main Lithuanian political parties - LVŽS and TS-LKD - for the 2016 and 2020 Seimas elections. The dataset consisted of around 300 paragraphs with populist text at a much higher frequency than "natural".

Using such a more "balanced" dataset would allow us to develop a model with an optimal balance between precision and recall characteristics. Key descriptive characteristics of the Train and Test Datasets are presented in Table 1. Detailed information on the Train and Test datasets can be found in GitHub.[3]

**Table 1. Train and Test Datasets**

|  | Train | Test |
|---|---|---|
| Texts | Specially collected "gold-standard" dataset | Lithuanian Manifestos* |
| Paragraph Count | 17 271 | 287 |
| AE Paragraphs | 497 | 130 |
| PC Paragraphs | 1 094 | 150 |

*Hand-coded Lithuanian party election manifestos for 2016 and 2020 elections

### 2.1.3. Hold-out dataset preparation

The hold-out dataset consists of the full-length manifestos of all political parties that stood for the 2016 and 2020 parliamentary election. Fourteen political parties ran for election in 2016 and 17 in 2020. The manifestos have been retrieved from the political parties' websites. Firstly, we have removed headings, tables and graphs from the manifestos and split the documents into paragraphs. Secondly, we coded the paragraphs have according to the above-described coding scheme. The paragraphs referring to "the people" as a single entity with homogeneous interests were coded as "people-centrist", and paragraphs criticizing the elite as a homogeneous unit have were coded as "anti-elitist". Then the coded paragraphs have been translated to English using "Google Translate". Finally, we double-checked the paragraphs where the results of hand-coding and machine coding differed. In many cases, the difference in coding appeared in borderline paragraphs, where it is complicated to establish if the paragraph contains populist content.

## 2.2. Model development

The development of machine-learning models can typically be broken down into three phases: data preparation/pre-processing, feature engineering and model training. The core methodology was to utilize a pre-trained BERT transformer model (Devlin et al. 2018) for feature engineering, which greatly simplified the data preparation and model training steps. This decision was motivated by the fact that similar approaches using BERT have already been successfully utilized for classifying political speech, e.g. classifying EU legislation to different topics (Chalkidis et al., 2019).

### 2.2.1. Data preparation

Typically, data preparation for machine-learning and natural language processing (NLP) tasks is a lengthy and complex process, which often feature such methodological steps as tokenization, stemming/lemming, dictionary-based encoding. However, as indicated earlier, the decision to use a pre-trained BERT model greatly simplified this stage. The version of the BERT model we used had an in-built tokenizer used for the initial model training. Therefore, the only pre-processing that was done was to break the text down into paragraphs (splitting on double newline characters "\n\n") and, where needed, translating texts to English. All the translations were done for individual paragraphs using "Google Translate".

---

[3] Project Repository: https://github.com/lukas-pkl/Populism/

### 2.2.2. Feature Engineering

The feature engineering step is crucial in any NLP project. In it, the natural text is converted into a numeric matrix format which is suitable for machine-learning models. Typically, with NLP, this involves some vectorization - a process where a unit of text (a single text, statement or a paragraph) is converted to a series of number - a vector. For example, the most straightforward text vectorization technique is one-hot encoding, where a dictionary is created from the unique words in the analyzed corpus, and each text is converted to a vector, based on which words from the dictionary appear in it. This technique has several drawbacks though very intuitive and straightforward - first, as the corpus grows, so does the dictionary and the length of the vector. For large and diverse corpora, the dictionaries and vectors can grow so large that computational analysis becomes complicated. Second, this technique considers each word individually, meaning that the more complex phrases and relationships between words are not captured. These drawbacks led to the development of "dense" vectorization techniques, where a text is converted to a fixed-length vector using more sophisticated algorithms (the most famous of which is 'word-2-vec' (Goldberg et al. 2014)). These techniques are attractive because they allow capturing the semantic meaning behind the texts to a certain degree (i.e. similar texts get similar vectors), and the result vectors stay the same size even as the analysis corpus grows. However, they also have a significant drawback - vectors produced in this way appear to be just arbitrary sequences of numbers, and the features encoded therein are not immediately recognizable.

Since 2017 such dense vectorization techniques developed rapidly, enabled by several innovations in the artificial neural networks, namely attention and transformer models (Vaswani et al. 2017). Arguably, the most famous and widely used of these models is BERT (Bidirectional Encoder Representations from Transformers). It is a large transformer model developed by researchers at Google (Devlin et al., 2018). The thing that makes this model particularly attractive is that a model pre-trained on the Google Books corpus has been shared publicly and is accessible to everyone. More specifically, it is possible to download and utilize a version of the BERT model, which produces a dense vector from any English text input.

We have used this BERT model (Variant: uncased_L-24_H-1024_A-16 (downloaded from Google Research GitHub Repo [https://github.com/google-research/bert109](https://github.com/google-research/bert109))) to vectorize the English texts paragraphs prepared in the previous step. BERT model turned these texts into numeric vectors with 1 024 dimensions. These dimensions constituted our feature sets and, with no additional augmentation, were used to develop the machine-learning models.

### 2.2.3. Model Training

We have developed several machine-learning models to generate the final predictions. All models were implemented using the "Scikit-Learn" (v 0.19.2) library in Python (Pedregosa et al., 2013). Since we have used an advanced transformer model in the feature engineering stage, we have aimed to keep this stage simple and use relatively simple machine-learning models with standard "out-of-the-box" parameters. Specifically, we developed the following models: Logistic Regression, Gaussian Naive Bayes, Support Vector Classifier, Multi-Layer Perceptron (MLP), and K Nearest Neighbors (KNN) Classifier. The performance of the models is presented in the table below. In the end, we have opted for an ensemble approach with the threshold of 2; i.e. a paragraph is considered either anti-elitist or people-centrist if two or more models produce a positive prediction. The results of the individual models and the model ensemble are presented in Table 2.

### 2.2.4. Model validation with the hold-out dataset

The developed models performed very well on a small test dataset. However, we sought to understand better how the model would perform on previously unseen data. To this end, we created an additional validation dataset and tested the model performance on it. As could be expected, the model performance metrics worsened on the validation dataset (see Table 3). The most significant deterioration was in the precision and F1 metrics for anti-elitism, meaning that the model generated more false positives than before. However, a manual check of these cases showed that a notable number of these paragraphs was borderline cases with some traces of "anti-elitism". This result illustrates that any numerical measure of a model's performance should be carefully considered when dealing with such "fuzzy" concepts with no clear boundaries: a high score does not in itself guarantee a good performance, and a low score does not necessarily mean the model is performing poorly. Overall, we did not observe any systematic bias and consider that the model performed reasonably well. A more detailed breakdown of the results is given in the next section.

**Table 2: ML Model Performance Using Test Dataset**

| Model | Accuracy (AE) | F1 (AE) | Precision (AE) | Recall (AE) | Accuracy (PC) | F1 (PC) | Precision (PC) | Recall (PC) |
|---|---|---|---|---|---|---|---|---|
| Logistic Regression | 0.63 | 0.33 | 0.96 | 0.2 | 0.66 | 0.52 | 0.96 | 0.36 |
| Gaussian Naïve Bayes | 0.64 | 0.71 | 0.56 | 0.95 | 0.59 | 0.72 | 0.57 | 0.97 |
| Support Vector Classifier | 0.86 | 0.84 | 0.87 | 0.82 | 0.76 | 0.78 | 0.72 | 0.87 |
| MLP | 0.71 | 0.59 | 0.83 | 0.45 | 0.7 | 0.66 | 0.84 | 0.54 |
| K-Nearest Neighbors Classifier | 0.59 | 0.72 | 0.79 | 0.03 | 0.59 | 0.2 | 0.74 | 0.13 |
| Ensemble (>=2) | 0.86 | 0.85 | 0.84 | 0.85 | 0.76 | 0.79 | 0.73 | 0.88 |

**Table 3: Ensemble Model Performance on the Hold-Out Dataset**

| Metric | AE | PC |
|---|---|---|
| Accuracy | 0.49 | 0.61 |
| F1 | 0.95 | 0.86 |
| Precision | 0.4 | 0.54 |
| Recall | 0.64 | 0.71 |

**3. Results**

This section of the paper presents the results of manual and model coding of manifestos of Lithuanian political parties that stood for the 2016 and 2020 parliamentary election. The percentage of people-centrist paragraphs ranges from less than 7% (LLP 2016, LSDP 2016)[4] to more than half of the manifesto (LLP 2020). The proportion of anti-elitist paragraphs in the manifestos is significantly lower - up to 1/3 in Taut2016.

---

[4] Abbreviations of the political parties' names are available in Appendix.

In general, machine-learning models predict populist paragraphs reasonably well, as in most cases, the proportion of manually and model coded paragraphs correspond.

As for inconsistencies, the model most often over-predicts anti-elitism in party manifestos. The LSDP (2020) manifesto has the largest difference of 9% between manual and model-coded paragraphs. This is followed by Taut (2016), LietuvaVisu (2020), LLRA (2020), APKK (2016). In terms of people-centrism, in some cases, the model over-predicts and, in other instances, under-predicts paragraphs in the manifestos. In the manifestos of LLRA (2016), KS (2020), Taut (2016) and TS-LKD (2016), the model significantly over-predicted the number of people-oriented paragraphs. For APKK (2016) and LLP (2020) manifestos, the model predicted fewer paragraphs than were coded manually. See Figure 1 and Figure 2.

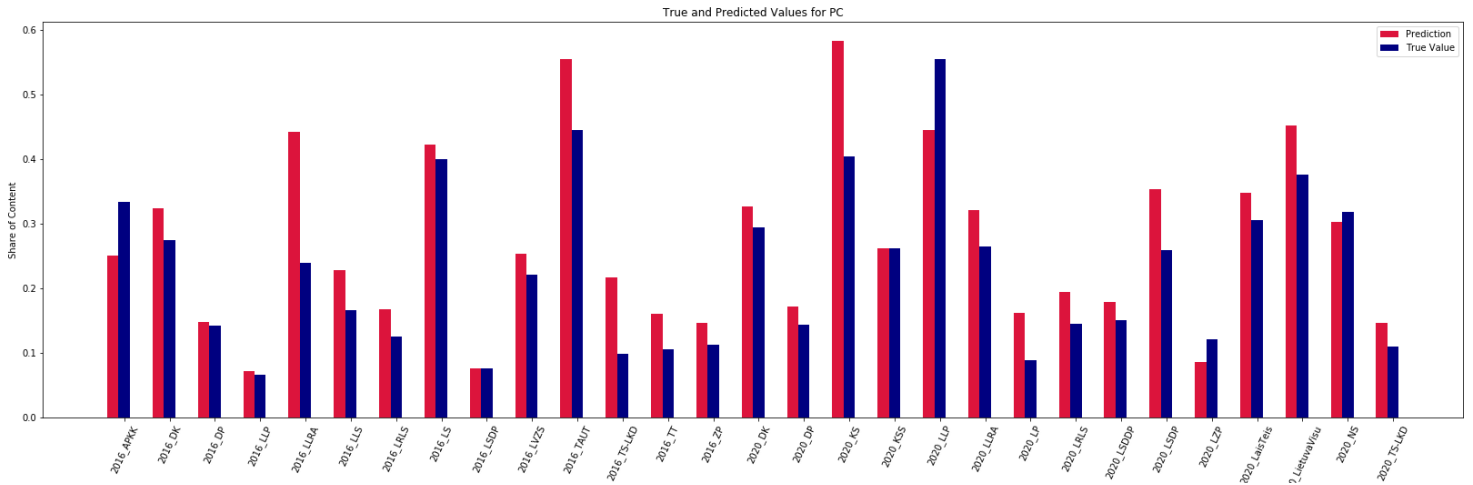

**Figure 1: True (blue) and Predicted (Red) values for People-Centrism by party**

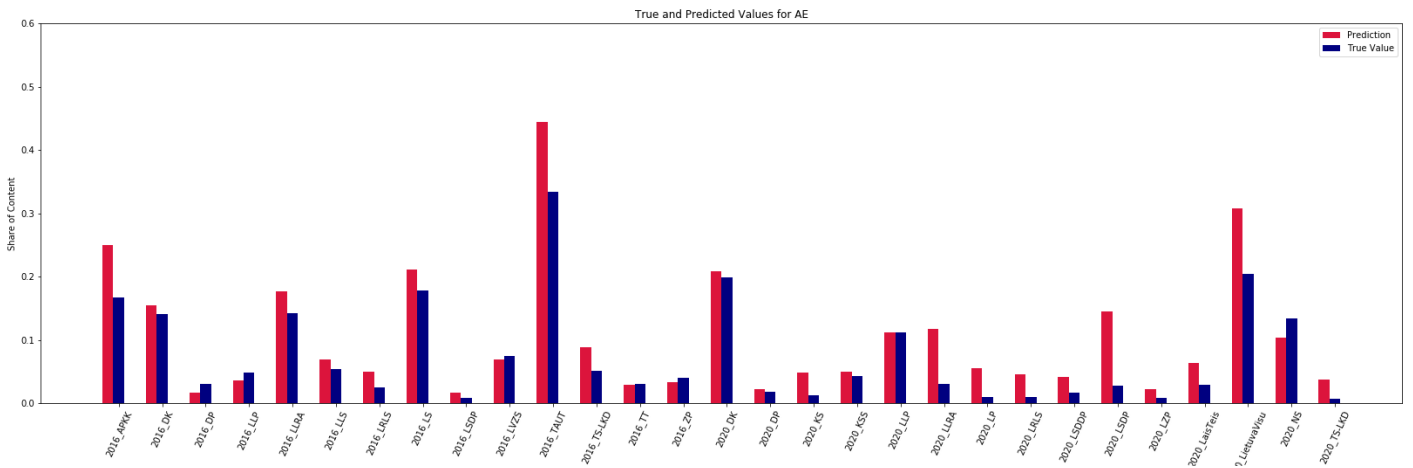

**Figure 2: True (blue) and Predicted (Red) values for Anti-Elitism by party**

The large difference between the hand-coding and machine coding of the Taut (2016) and APKK (2016) manifestos could be due to the very low number of paragraphs (see Annex), as each incorrectly coded paragraph has a significant impact on the proportion of populist paragraphs in the manifestos. In addition, the over-prediction of populist paragraphs by machine learning models is because we have chosen the coding correspondence of two models as the threshold to consider a paragraph to be populist or anti-elitist. If this threshold increased (three-model agreement), the number of paragraphs would be lower. However, we preferred models that over-predicted rather than under-predicted populism in manifestos since ML model coding combination with human verification could lead to better final results.

**Summary**

We conclude that machine-learning models predict populist paragraphs reasonably well. Given the indeterminacy of the phenomenon, context-dependence, vagueness of the key concepts (the elite and the people) and the complexity of their operationalization, we conclude that machine-learning models perform coding relatively successfully.

- It is possible to develop ML model to identify populist content in text;
- This model can work well in different contexts and can handle new-unseen data well;
- For best results, keep a human in the loop and validate model predictions

The model developed here is not sufficiently well-performing to be considered a fully-automated solution, but it can be used as a tool to assist human coders by producing a short-list of paragraphs which can then be verified by a human expert (this is often called a "human-in-the-loop" system). As the model produces relatively few false-negatives, such human in loop system has the potential to significantly reduce the amount of work needed to code vast amounts of texts without any loss of quality.

# Appendix

Table 1: List of manifestos of Lithuanian political parties

|   | Party name (LT) | Party name (EN) | Abbreviation |
|---|---|---|---|
|   | **Parliamentary election 2016** |   |   |
| 1 | Darbo partija | Labour Party | DP |
| 2 | „Drąsos kelias" politinė partija | Political party 'The Way of Courage' | DK |
| 3 | Antikorupcinė N. Puteikio ir K. Krivicko koalicija | Anti-Corruption Coalition of Kristupas Krivickas and Naglis Puteikis | APKK |
| 4 | Lietuvos laisvės sąjunga (liberalai) | Lithuanian Freedom Union (Liberals) | LLS |
| 5 | Lietuvos lenkų rinkimų akcija-Krikščioniškų šeimų sąjunga | Electoral Action of Poles in Lithuania – Christian Families Alliance | LLRA |
| 6 | Lietuvos liaudies partija | Lithuanian People's Party | LLP |
| 7 | Lietuvos Respublikos liberalų sąjūdis | Liberal Movement of the Republic of Lithuania | LRLS |
| 8 | Lietuvos socialdemokratų partija | Social Democratic Party of Lithuania | LSDP |
| 9 | Lietuvos valstiečių ir žaliųjų sąjunga | Lithuanian Farmers and Greens Union | LVŽS |
| 10 | Lietuvos žaliųjų partija | Lithuanian Green Party | LŽP |
| 11 | Partija Tvarka ir teisingumas | Party "Order and Justice" | TT |
| 12 | Politinė partija „Lietuvos sąrašas" | Political party 'Lithuanian List' | LS |
| 13 | Tėvynės sąjunga - Lietuvos krikščionys demokratai | Homeland Union – Lithuanian Christian Democrats | TS-LKD |
| 14 | Tautininkų sąjunga | S. Buškevičius and Nationalists' Coalition 'Against the Corruption and Poverty' | Taut |
|   | Parliamentary election 2020 |   |   |
| 1 | Darbo partija | Labour Party | DP |
| 2 | „Drąsos kelias" politinė partija | Political party 'The Way of Courage' | DK |
| 3 | KARTŲ SOLIDARUMO SĄJUNGA - SANTALKA LIETUVAI | Union of Intergenerational Solidarity – Cohesion for Lithuania | KSS |
| 4 | Krikščionių sąjunga | Christian Union | KS |
| 5 | Laisvės partija | Freedom Party | LP |
| 6 | Lietuvos lenkų rinkimų akcija - Krikščioniškų šeimų sąjunga | Electoral Action of Poles in Lithuania – Christian Families Alliance | LLRA |
| 7 | Lietuvos liaudies partija | Lithuanian People's Party | LLP |
| 8 | Lietuvos Respublikos liberalų sąjūdis | Liberal Movement of the Republic of Lithuania | LRLS |
| 9 | Lietuvos socialdemokratų darbo partija | Social Democratic Labour Party of Lithuania | LSDDP |
| 10 | Lietuvos socialdemokratų partija | Social Democratic Party of Lithuania | LSDP |
| 11 | Lietuvos valstiečių ir žaliųjų sąjunga | Lithuanian Farmers and Greens Union | LVŽS |
| 12 | Lietuvos žaliųjų partija | Lithuanian Green Party | LŽP |
| 13 | Nacionalinis susivienijimas | National Alliance | NS |
| 14 | Partija „Laisvė ir teisingumas" | Party "Freedom and Justice" | LaisTeis |
| 15 | Partija LIETUVA – VISŲ | Party "Lithuania – For everyone" | LietuvaVisu |
| 16 | Politinė partija „Lietuvos sąrašas" | Political party 'Lithuanian List' | LS |
| 17 | Tėvynės sąjunga – Lietuvos krikščionys demokratai | Homeland Union – Lithuanian Christian Democrats | TS-LKD |

Table 2: Proportion of populist paragraphs
in hold-out dataset

| Year | Party | AE-True | AE-Pred | PC-True | PC2-Pred | F1-AE | F1-PC | Paragraph Count |
|---|---|---|---|---|---|---|---|---|
| 2016 | APKK | 0,25 | 0,17 | 0,25 | 0,33 | 0,80 | 0,57 | 11 |
| 2016 | DK | 0,15 | 0,14 | 0,32 | 0,27 | 0,71 | 0,68 | 141 |
| 2016 | DP | 0,02 | 0,03 | 0,15 | 0,14 | 0,34 | 0,65 | 622 |
| 2016 | LLP | 0,04 | 0,05 | 0,07 | 0,07 | 0,57 | 0,26 | 166 |
| 2016 | LLRA | 0,18 | 0,14 | 0,44 | 0,24 | 0,72 | 0,65 | 112 |
| 2016 | LLS | 0,07 | 0,05 | 0,23 | 0,17 | 0,54 | 0,66 | 390 |
| 2016 | LRLS | 0,05 | 0,03 | 0,17 | 0,12 | 0,39 | 0,61 | 1104 |
| 2016 | LS | 0,21 | 0,18 | 0,42 | 0,40 | 0,86 | 0,86 | 89 |
| 2016 | LSDP | 0,02 | 0,01 | 0,08 | 0,08 | 0,50 | 0,69 | 473 |
| 2016 | LVZS | 0,07 | 0,07 | 0,25 | 0,22 | 0,59 | 0,67 | 1019 |
| 2016 | TAUT | 0,44 | 0,33 | 0,56 | 0,44 | 0,57 | 0,89 | 8 |
| 2016 | TS-LKD | 0,09 | 0,05 | 0,22 | 0,10 | 0,46 | 0,45 | 1863 |
| 2016 | TT | 0,03 | 0,03 | 0,16 | 0,11 | 0,34 | 0,52 | 491 |
| 2016 | LZP | 0,03 | 0,04 | 0,15 | 0,11 | 0,31 | 0,65 | 177 |
| 2020 | DK | 0,21 | 0,20 | 0,33 | 0,29 | 0,81 | 0,76 | 210 |
| 2020 | DP | 0,02 | 0,02 | 0,17 | 0,14 | 0,52 | 0,64 | 658 |
| 2020 | KS | 0,05 | 0,01 | 0,58 | 0,40 | 0,00 | 0,72 | 83 |
| 2020 | KSS | 0,05 | 0,04 | 0,26 | 0,26 | 0,29 | 0,71 | 305 |
| 2020 | LLP | 0,11 | 0,11 | 0,44 | 0,56 | 1,00 | 0,89 | 17 |
| 2020 | LLRA | 0,12 | 0,03 | 0,32 | 0,27 | 0,42 | 0,82 | 163 |
| 2020 | LP | 0,06 | 0,01 | 0,16 | 0,09 | 0,20 | 0,45 | 1078 |
| 2020 | LRLS | 0,05 | 0,01 | 0,19 | 0,14 | 0,29 | 0,60 | 1358 |
| 2020 | LSDDP | 0,04 | 0,02 | 0,18 | 0,15 | 0,56 | 0,62 | 429 |
| 2020 | LSDP | 0,15 | 0,03 | 0,35 | 0,26 | 0,28 | 0,58 | 330 |
| 2020 | LZP | 0,02 | 0,01 | 0,09 | 0,12 | 0,36 | 0,53 | 362 |
| 2020 | LaisTeis | 0,06 | 0,03 | 0,35 | 0,30 | 0,31 | 0,74 | 140 |
| 2020 | LietuvaVisu | 0,31 | 0,20 | 0,45 | 0,38 | 0,64 | 0,79 | 220 |
| 2020 | NS | 0,10 | 0,13 | 0,30 | 0,32 | 0,58 | 0,70 | 260 |
| 2020 | TS-LKD | 0,04 | 0,01 | 0,15 | 0,11 | 0,16 | 0,52 | 1403 |

Table 3: Numbers of coded paragraphs in hold-out dataset

| Year | Party | AE Paragraphs | PC Paragraphs | Paragraphs Total |
|---|---|---|---|---|
| 2016 | APKK | 3 | 3 | 11 |
| 2016 | DK | 22 | 46 | 141 |
| 2016 | DP | 10 | 92 | 622 |
| 2016 | LLP | 6 | 12 | 166 |
| 2016 | LLRA | 20 | 50 | 112 |
| 2016 | LLS | 27 | 89 | 390 |
| 2016 | LRLS | 55 | 185 | 1104 |
| 2016 | LS | 19 | 38 | 89 |
| 2016 | LSDP | 8 | 36 | 473 |
| 2016 | LVZS | 70 | 258 | 1019 |
| 2016 | TAUT | 4 | 5 | 8 |
| 2016 | TS-LKD | 165 | 405 | 1863 |
| 2016 | TT | 14 | 79 | 491 |
| 2016 | LZP | 6 | 26 | 177 |
| 2020 | DK | 44 | 69 | 210 |
| 2020 | DP | 15 | 113 | 658 |
| 2020 | KS | 4 | 49 | 83 |
| 2020 | KSS | 15 | 80 | 305 |
| 2020 | LLP | 2 | 8 | 17 |
| 2020 | LLRA | 19 | 52 | 163 |
| 2020 | LP | 60 | 174 | 1078 |
| 2020 | LRLS | 62 | 265 | 1358 |
| 2020 | LSDDP | 18 | 77 | 429 |
| 2020 | LSDP | 48 | 117 | 330 |
| 2020 | LZP | 8 | 31 | 362 |
| 2020 | LaisTeis | 9 | 49 | 140 |
| 2020 | LietuvaVisu | 68 | 100 | 220 |
| 2020 | NS | 27 | 79 | 260 |
| 2020 | TS-LKD | 53 | 206 | 1403 |